\documentclass[lettersize,journal]{IEEEtran}
\usepackage{amsmath,amsfonts}
\usepackage{algorithmic}
\usepackage{algorithm}
\usepackage{array}
\usepackage[caption=false,font=normalsize,labelfont=sf,textfont=sf]{subfig}
\usepackage{times}
\usepackage{epsfig}
\usepackage{amsmath}
\usepackage{ragged2e}
\usepackage{textcomp}
\usepackage{stfloats}
\usepackage{url}
\usepackage{verbatim}
\usepackage{graphicx}
\usepackage{cite}
\usepackage{multirow}
\usepackage{tabularx}
\usepackage{booktabs}
\usepackage{color}
\usepackage{hyperref}
\ifdefined \GramaCheck
\newcommand{\CheckRmv}[1]{}
\newcommand{\figref}[1]{Figure 1}%
\newcommand{\tabref}[1]{Table 1}%
\newcommand{\secref}[1]{Section 1}
\renewcommand{\eqref}[1]{Equation 1}
\else
\newcommand{\CheckRmv}[1]{#1}
\newcommand{\figref}[1]{Fig.~\ref{#1}}%
\newcommand{\tabref}[1]{Table~\ref{#1}}%
\newcommand{\secref}[1]{Sec.~\ref{#1}}
\renewcommand{\eqref}[1]{Equation~(\ref{#1})}
\fi

\newcommand{\our}{{C$^2$F-Net}}
\newcommand{\ourM}{{C$^2$F-Net}}
\newcommand{\ourB}{CIM} 

\def\ie{\emph{i.e.}}
\def\eg{\emph{e.g.}}
\def\etc{\emph{etc}}
\def\etal{{\em et al.}}

\def\etal{{\em et al.}}

\usepackage[normalem]{ulem}
\graphicspath{{./figures/}{./authors/}}
\begin{document}
\title{Camouflaged Object Detection via Context-aware\\ Cross-level Fusion}

\author{Geng Chen$^{*}$,~
	Si-Jie Liu$^{*}$,~
	Yu-Jia Sun,~
	Ge-Peng Ji,~
	Ya-Feng Wu,~
	Tao Zhou, \IEEEmembership{Member, IEEE}

\thanks{G. Chen, S.-J. Liu, and Y.-F. Wu are with Northwestern Polytechnical University, Xi'an 710072, China. G. Chen is also with National Engineering Laboratory for Integrated Aero-Space-Ground-Ocean Big Data Application Technology, School of Computer Science and Engineering. S.-J. Liu and Y.-F. Wu are also with Data Processing Center, School of Power and Energy (Emails: geng.chen.cs@gmail.com; sijieliu\_123@sina.com; yfwu@nwpu.edu.cn).}
\thanks{T. Zhou is with Key Laboratory of System Control and Information Processing, Ministry of Education, Shanghai 200240, China, and School of Computer Science and Engineering, Nanjing University of Science and Technology, Nanjing 210094, China (Email: taozhou.ai@gmail.com).}
\thanks{Y.-J. Sun is with School of Computer Science, Inner Mongolia University, Hohhot 010021, China (Email: thograce@163.com).}
\thanks{G.-P. Ji is with Artificial Intelligence Institute, School of Computer Science, Wuhan University, Wuhan 430072, China (Email: gepengai.ji@gmail.com).}

\thanks{Equal contribution: Geng Chen and Sijie Liu.}
\thanks{Corresponding authors: Tao Zhou and Yafeng Wu.}
\thanks{Copyright © 2022 IEEE. Personal use of this material is permitted.
However, permission to use this material for any other purposes must be obtained from the IEEE by sending an email to pubs-permissions@ieee.org.}}

\markboth{Journal of \LaTeX\ Class Files,~Vol.~14, No.~8, August~2021}%
{Chen  \MakeLowercase{\textit{et al.}}: Camouflaged Object Detection via Context-aware Cross-level Fusion}

\maketitle
\IEEEdisplaynontitleabstractindextext
\IEEEpeerreviewmaketitle
\begin{abstract}
Camouflaged object detection (COD) aims to identify the objects that conceal themselves in natural scenes. Accurate COD suffers from a number of challenges associated with low boundary contrast and the large variation of object appearances,~\eg, object size and shape.
To address these challenges, we propose a novel Context-aware Cross-level Fusion Network (\ourM), which fuses context-aware cross-level features for accurately identifying camouflaged objects.
Specifically, we compute informative attention coefficients from multi-level features with our Attention-induced Cross-level Fusion Module (ACFM), which further integrates the features under the guidance of attention coefficients.
We then propose a Dual-branch Global Context Module (DGCM) to refine the fused features for informative feature representations by exploiting rich global context information. Multiple ACFMs and DGCMs are integrated in a cascaded manner for generating a coarse prediction from high-level features.
The coarse prediction acts as an attention map to refine the low-level features before passing them to our Camouflage Inference Module (CIM) to generate the final prediction.
We perform extensive experiments on three widely used benchmark datasets and compare \ourM~with state-of-the-art (SOTA) models. The results show that \ourM~is an effective COD model and outperforms SOTA models remarkably. Further, an evaluation on polyp segmentation datasets demonstrates the promising potentials of our \ourM~in COD downstream applications.
Our code is publicly available at: \url{https://github.com/Ben57882/C2FNet-TSCVT}.
\end{abstract}


\section{Introduction}\label{sec:introduction}

\IEEEPARstart{O}{ver} the course of evolution, wild animals have developed extensive camouflage abilities in order to survive.
In practice, they try to alter their appearance to ``perfectly'' blend in with their surroundings in order to avoid attracting the notice of other creatures.
Recently, camouflage has attracted increasing research interest from the computer vision community \cite{anet,sinet,rethinking}.
Among various topics, camouflaged object detection (COD), which aims to identify and segment camouflaged objects from images, is particularly popular.
However, accurate COD is with considerable difficulties due to the special characteristics of camouflage.
More specifically, due to the camouflage, the boundary contrast between an object and its surroundings is extremely low, leading to significant difficulties to identify/segment this object.
As shown in the top row of \figref{fig:examples}, it is very challenging to discover {the butterfly from the leaves.
Also, in the middle row of \figref{fig:examples}, it is a tremendous task to find out the tortoise occluded by leaves on the ground.}
In addition, the camouflaged objects, most wild animals, are usually with varied appearances, \eg, size and shape, which further aggravates the difficulties in accurate COD.
\CheckRmv{
	\begin{figure}[t]
		\centering
		\includegraphics[width=1\linewidth]{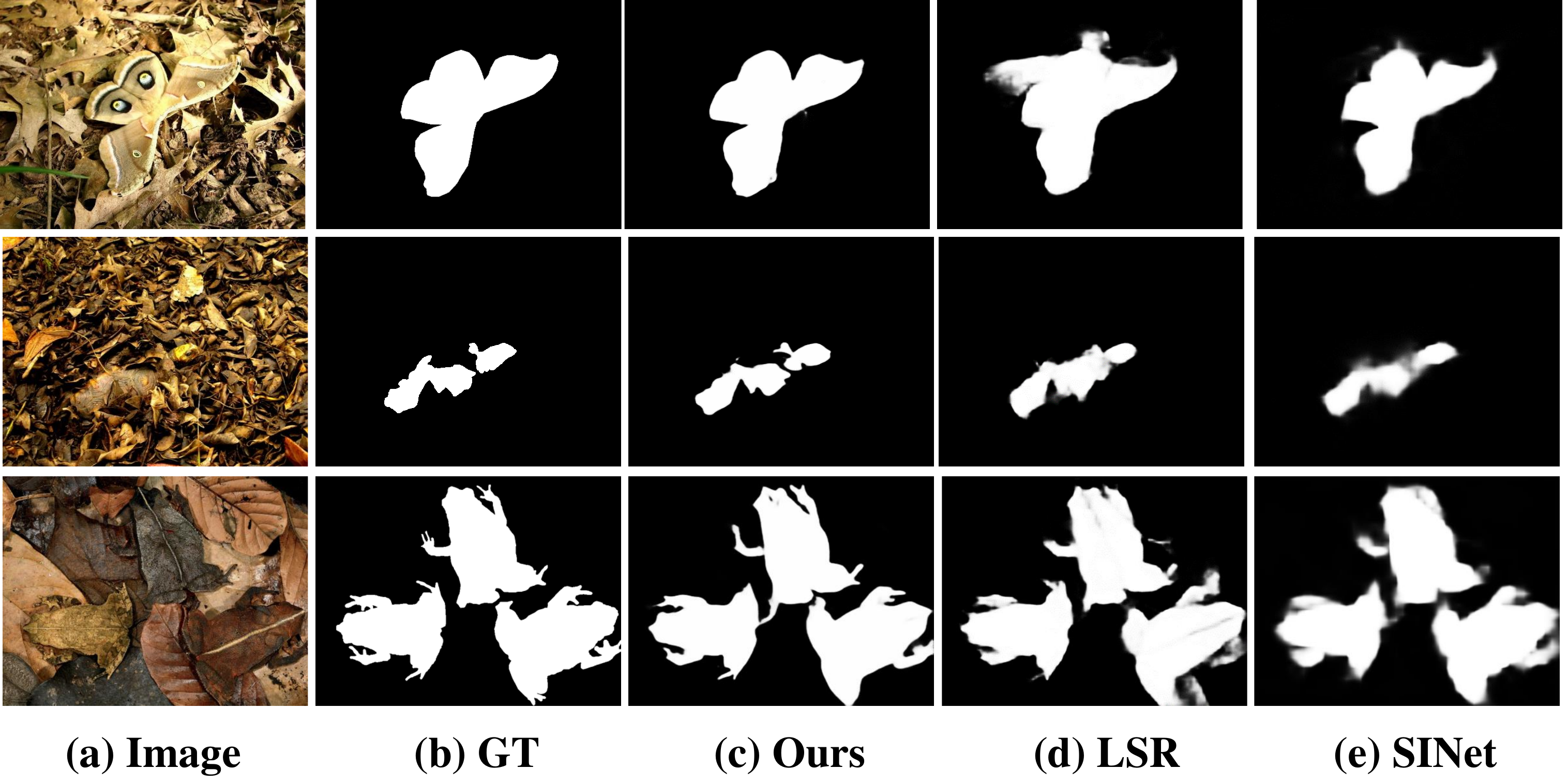}
		\vspace{-5pt}
		\caption{We show three challenging camouflage scenarios (top to bottom) with indefinable boundary, occluded object, and multiple objects. Our model outperforms cutting-edge models, 
			SINet \protect\cite{sinet} and LSR\protect\cite{lsr}, in these challenging scenarios.}
		\label{fig:examples}
	\end{figure}
}

To tackle these challenges, deep learning techniques~\cite{cubenet} have been adopted and shown great potential.
In addition, a number of COD datasets have been constructed for training the deep learning models.
For instance, Le \etal~\cite{anet} created the first COD dataset, called CAMO, consisting of 2,500 images.
However, CAMO, which suffers from a limited sample size, is insufficient for taking full advantage of deep learning models.
Recently, Fan \etal~\cite{sinet} constructed COD10K, the first large-scale COD dataset consisting of 10,000 images with the consideration of various challenging camouflage attributes in real natural environments.
In addition to datasets, these two works also contribute to the COD from the aspect of deep learning models. Le \etal~\cite{anet} proposed an anabranch network, which classifies whether an image contains camouflaged objects and then integrates this information into the COD task. Motivated by the fact that predators discover preys by searching first and then recognizing, Fan \etal~\cite{sinet} designed searching modules to identify the rough areas of camouflaged objects and then accurately segmented camouflaged objects with identification modules.

Although existing methods have shown promising performance in the detection of a single camouflaged object from relatively simple scenes, their performance degrades for a number of challenging cases, \eg, {occlusion, multiple objects}, etc.
As shown in the top two rows of \figref{fig:examples}, state-of-the-art models show unsatisfactory performance when occlusion exits or the boundary of a camouflaged object is indefinable. In addition, they are unable to accurately identify the boundaries of camouflaged objects when multiple camouflaged objects exist, as shown in the bottom row of \figref{fig:examples}.
These challenges can be addressed with a significantly large receptive field, which provides rich context information for accurate COD.
In addition, how to effectively fuse cross-level features also plays a crucial role in the success of COD.
However, existing works usually overlook the importance of both of these two key factors. It is, therefore, greatly desired for a unified COD framework that jointly considers rich context information and effective cross-level feature fusion.

To this end, we propose Context-aware Cross-level Fusion Network (\ourM), a novel deep learning model for accurate COD.
In \ourM, the cross-level features extracted from the backbone are first fused with an attention-induced cross-level fusion module (ACFM), which achieves the feature integration with the clues from a multi-scale channel attention (MSCA) component.
More specifically, the ACFM involves three major steps, including (i) attention coefficient computation from multi-level features, (ii) feature refinement with the attention coefficients, and (iii) feature integration for the fused one.
Subsequently, a dual-branch global context module (DGCM) is proposed to exploit the rich context information from the fused features.
The DGCM transforms the input features into multi-scaled ones with two parallel branches, employs the MSCA components to compute the attention coefficients, and integrates the features under the guidance of the attention coefficients.
Multiple ACFMs and DGCMs are organized in a cascaded manner at two stages, from high-level to low-level.
The final DGCM predicts a coarse segmentation map of the camouflaged object(s).
We then refine the low-level features with the obtained coarse map and perform the final detection with the proposed Camouflage Inference Module (CIM).
In summary, our contributions are as follows:
\begin{itemize}
	\item We propose \ourM, a novel COD model that integrates the cross-level features by considering rich global context information, effective cross-level fusion, and low-level feature refinement.
	\item We propose DGCM, a context-aware module that can capture valuable context information for improving the accuracy of COD. Furthermore, we integrate the cross-level features with ACFM, a novel fusion module that integrates the features with the valuable attention cues provided by MSCA.
	\item We propose CIM, an effective module capable of exploiting informative features from the low-level features refined by a coarse prediction map.
	\item Extensive experiments on three benchmark datasets demonstrate that our \ourM~outperforms 19 state-of-the-art models in the terms of five evaluation metrics for the COD task.
	Further poly segmentation experiments indicate the superior performance of our \ourM~in COD downstream applications.
\end{itemize}

This paper significantly extends our previous work published in the IJCAI 2021 \cite{c2f}, with multi-fold improvements as follows.
(i) We improve the model by refining the low-level features with the coarse COD map and then predicting the final result with our CIM. The new improvement is detailed in Sec.~\ref{cim} and has shown promising performance in both quantitative and qualitative evaluations (see Sec.~\ref{sec:experiments}).
(ii) We provide more details to the conference version. Specifically, we add a subsection to review the downstream applications of COD and present recent advances in poly segmentation (see Sec.~\ref{applications}). We also provide the details of evaluation metrics to better understand their characteristics (see Sec.~\ref{metrics}). 
(iii) We compare the proposed model with more state-of-the-art methods to validate the effectiveness of our model (see Sec.~\ref{comparison}), and provide a sub-class comparison and a more insightful discussion (see Sec.~\ref{subclass}).
(iv) We extend the proposed \ourM~to a downstream application of COD, \ie, polyp segmentation. Quantitative and qualitative evaluations conducted on four benchmark datasets demonstrate the superiority of our model over other existing polyp segmentation methods (see Sec.~\ref{polyseg}).

The rest of this paper is arranged as follows. In Section~\ref{related_work}, we review a number of works that are closely related to ours. In Section~\ref{Methods}, we describe our \ourM~in detail. In Section~\ref{sec:experiments}, we provide extensive experimental results, ablation studies, and the further application of our \ourM~in polyp segmentation. Finally, we conclude this work in Section~\ref{conclusion}.

\section{Related Work} \label{related_work}

In this section, we discuss three types of works that are closely related to our method, \ie, camouflaged object detection, context-aware deep learning, and downstream applications of COD.

\subsection{Camouflaged Object Detection}

Recently, significant efforts have been put into COD, which is an emerging field in the computer vision community.
Early COD approaches identify camouflaged objects with visual features, including texture, color, gradient, and motion~\cite{survey}.
In practice, a single visual feature cannot provide comprehensive characteristics for camouflaged objects.
Therefore, multiple features are integrated for improving performance \cite{comb1}.
Moreover, the Bayesian framework has been employed for detecting moving camouflaged objects from videos \cite{Bayesian}.
Despite their advantages, existing approaches relying on hand-crafted features tend to fail in real-world applications since they can only work with relatively simple scenarios. 
To this end, deep learning models, which automatically learn features and are trained in an end-to-end manner, have been adopted for accurate COD, which acts as an effective solution to address the challenges associated with traditional hand-crafted features.
Yan \etal~\cite{mirrornet} proposed a two-stream network, called MirrorNet, for COD with the original and flipped images. The underlying motivation lies in the fact that the flipped image can provide valuable cues for COD.
Lamdouar \etal~\cite{video} identified camouflaged objects from videos by exploiting the motion information with a deep learning framework, which consists of two modules, \ie, a differentiable registration module and a motion segmentation module. 
Ren~\etal~\cite{TANet} proposed a two-step texture-aware refinement module to amplify the differences between the camouflaged object and its surroundings for accurate COD.
Ji~\etal~\cite{errnet} designed a multivariate calibration strategy to contrast the initial edge prior obtained from a selective edge aggregation strategy that mimics human visual perception.
Li~\etal~\cite{li2021uncertainty} presented an adversarial learning network with a similarity measure module to model the contradictory information, enhancing the ability to detect salient and camouflaged objects.
Mei~\etal~\cite{pfnet} developed a bio-inspired framework named Positioning and Focus Network, which contains a positioning module for mimicking the detection process of predator and a focus module to perform the identification process.
Lyu~\etal~\cite{lsr} proposed the first ranking-based network based on Mask-RCNN~\cite{maskrcnn} to learn the camouflage degree with joint fixation and camouflage decoder. They also collect the large-scale testing dataset NC4K.
SINetV2~\cite{fan2021concealed} designed a neighbor connection decoder and group-reversal attention for detecting camouflaged objects.
Interested readers can refer to \cite{rethinking} for a comprehensive review of COD.
Furthermore, it is worth noting that COD shares a number of similarities with the popular salient object detection (SOD) tasks, such as RGB SOD~\cite{egnet}, RGB-Depth/RGB-Thermal SOD \cite{zhang2019rgb,chen2020dpanet,huang2021employing}, remote sensing SOD \cite{cong2021rrnet,zhang2020dense}, Co-SOD \cite{cong2017iterative,fan2021rethinking}, etc. 
Different from existing models, our \ourM~advances the COD accuracy by jointly considering the rich global context information, effective cross-level fusion, and low-level feature refinement.
\subsection{Context-aware Deep Learning}
Due to its superior capability of enhancing the feature representation, contextual information acts as a vital role in segmenting objects.
For this reason, efforts have been made to enrich the contextual information.
For instance, Zhao \etal~\cite{pspnet} proposed PSPNet to establish multi-scale representations around each pixel for obtaining rich context information.
Chen \etal~\cite{deeplab} constructed ASPP with different dilated convolutions in order to enrich context information.
The self-attention mechanism has also been proposed and employed for capturing rich context information. Typical instances include DANet~\cite{danet}, which extracts contextual information with non-local modules~\cite{nonlocal}, and CCNet~\cite{ccnet}, which employs multiple cascaded cross-attention modules to obtain dense contextual information.
Context information also gains significant attention from the field of object segmentation. For instance, Zhang \etal~\cite{bi} enriched context information with a multi-scale context-aware feature extraction module, which provides rich context information.
Liu \etal~\cite{poolnet} proposed PoolNet, which is equipped with a pyramid pool module for rich context features highly related to the accurate salient object detection.
Chen \etal~\cite{gcpanet} proposed a global context flow module to transfer features containing global semantic information to multi-level features at different stages to improve the integrity of SOD.
Tan \etal~\cite{lcanet} integrated the local and global context features in an adaptive coarse-to-fine manner for improved feature representations. Besides, the motion context and information have been effectively exploited and used in motion prediction \cite{shu2021spatiotemporal} and activity recognition \cite{shu2022expansion}. 
 
\subsection{Downstream Applications of COD}\label{applications}

COD has a large number of downstream applications in practice. Typical instances include medical image segmentation (\eg, polyp segmentation~\cite{ji2022vps,ji2021pnsnet}, COVID-19 lung infection segmentation~\cite{fan2020infnet}, \etc.), surface defect detection, recreational art, concealed-salient object transition, transparent objects detection, search engines, \etc \cite{fan2021concealed}.

Among various applications, polyp segmentation is one of the most attractive tasks due to its vital application value and close relationship with COD.
Accurate polyp segmentation can provide vital information from colonoscopy images for identifying polyps, which have a high risk to develop into serious colorectal cancer that threatens human life.
Early polyp segmentation methods rely on hand-crafted features \cite{mamonov2014automated}, which are unable to fully capture the useful information for segmenting polyps.
To address these challenges, deep learning has been employed for polyp segmentation and has shown great success.
For instance, Brandao \etal \cite{brandao2017fully} utilized the fully convolutional networks (FCNs) to segment polyps in an end-to-end manner.
Akbari \etal \cite{akbari2018polyp} employed modified FCNs for accurate polyp segmentation.
The variants of U-Net, \ie, U-Net++~\cite{unetplus} and ResUNet++~\cite{jha2019resunetplus}, were also utilized for polyp segmentation for improved performance.
Boundary information was also employed for improving the accuracy of polyp segmentation \cite{murugesan2019psi,fang2019selective}.
Ji \etal \cite{ji2021pnsnet} proposed a normalized self-attention module to progressively segment the polyp region in the video clips. 
More recently, Fan \etal \cite{pranet} proposed PraNet to segment polyps with paralleled partial decoder and reverse attention mechanism, significantly improving the performance.
In this study, we further demonstrate the effectiveness of our \ourM~by applying it to the polyp segmentation task.

\begin{figure*}[ht]
	\centering{
	}
	\includegraphics[width=\textwidth]{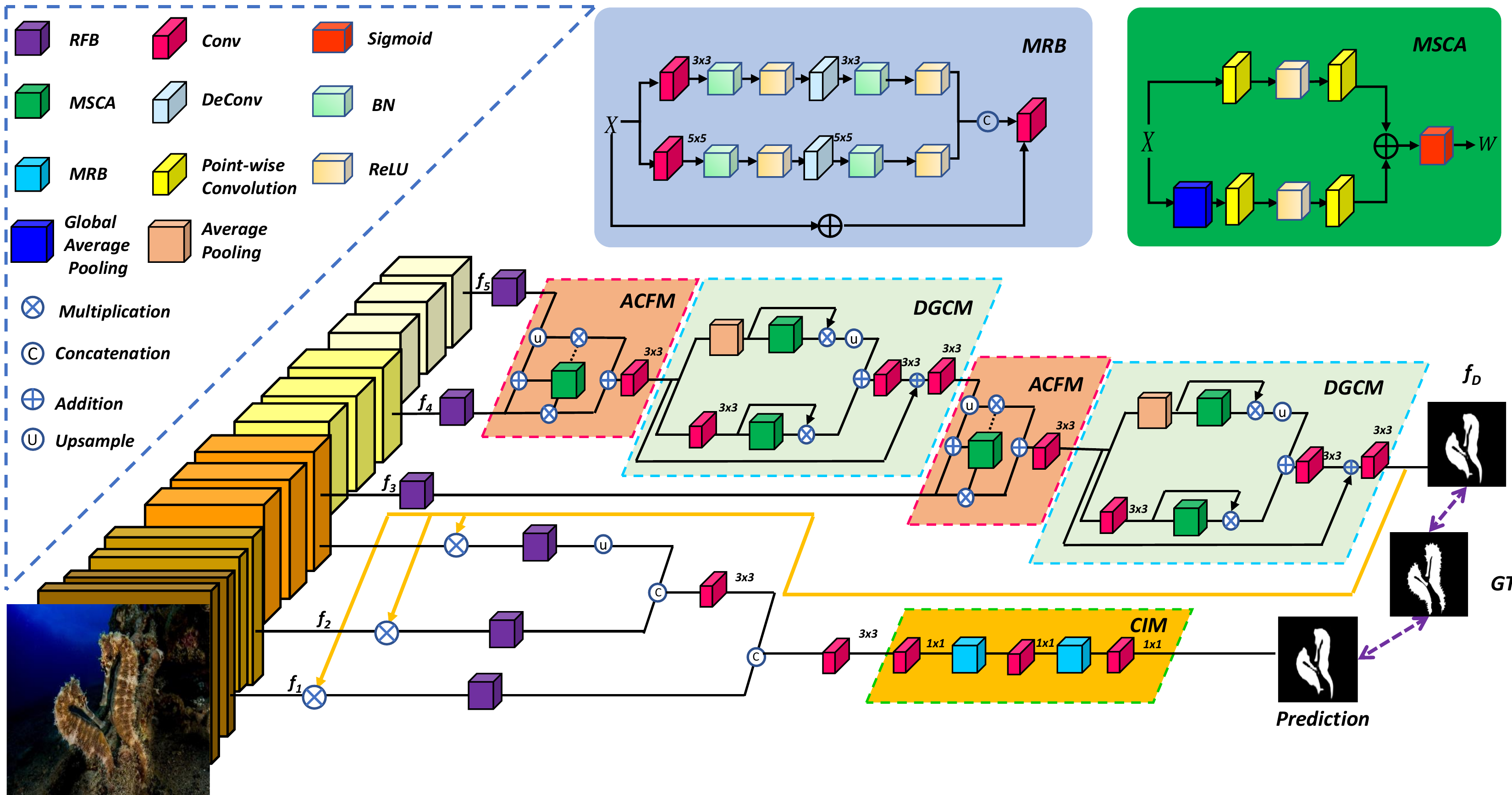}
	\caption{The overall framework of our \our, including three main components, \ie, attention-induced cross-level fusion module (ACFM), dual-branch global context module (DGCM), and~camouflage inference module (CIM). Note that MRB and MSCA are short names for multi-scale residual block and multi-scale channel attention, respectively.
	See Sec.~\ref{Methods} for details.
	}
	\label{fig:net1}
\end{figure*}

\section{Proposed Method}\label{Methods}

In this section, we first present the overall framework of the proposed \ourM. Then, we detail three key components in our \ourM. Finally, we detail the overall loss function for the proposed COD method.

\subsection{Overall Architecture} 

\figref{fig:net1} shows the overall framework of our \ourM, which can effectively integrate context-aware cross-level features to boost the accuracy of camouflaged object detection. Specifically,  Res2Net-50~\cite{res2net} is adopted as the backbone to extract multi-scale features, denoted as $f_i$ with $i=1,2,\dots,5$.
In our model, the extracted features are divided into two groups, \ie, low-level features $Q_l = \{f_1,f_2,f_3\}$ and high-level features $Q_h = \{f_3,f_4,f_5\}$.
Then, three receptive field blocks (RFB) are utilized to enrich the features from $Q_h$ by expanding the receptive fields. We utilize the same setting in \cite{res2net}, and the RFB includes five branches $b_k$ with $k=1,2,\dots,5$. In each branch, a $1\times{1}$ convolutional layer is first adopted to reduce the original channel size to 64. Following that, a $(2k-1)\times(2k-1)$ convolutional layer and a $3\times{3}$ convolutional layer with a specific dilation rate $(2k-1)$ (when $k>2$) are adopted. The first four branches are concatenated to obtain the fused feature, which is fed into a $1\times{1}$ convolutional layer for reducing the channel size to $64$. Besides, the output of the fifth branch is added by the feature from the first four branches, which is further fed to a \emph{ReLU} activation to obtain the final feature. Moreover, an attention-induced cross-level fusion module (ACFM) is proposed to fuse high-level features, and a dual-branch global context module (DGCM) is proposed to fully exploit multi-scale context information from the fused features. Two modules are organized in a cascaded manner. The last DGCM provides an initial prediction $f_D$. We then refine the low-level features $Q_l$ with $f_D$ and predict the final COD result with our camouflage inference module (CIM).
\subsection{Attention-induced Cross-level Fusion Module}

Natural differences often exist among different types of camouflaged objects. What's more, due to the observation distance and relative locations to the surrounding background, the size of similar camouflaged objects could change dramatically. To address the above issues, we design a novel ACFM by introducing a multi-scale channel attention (MSCA)~\cite{aff} strategy to integrate cross-level features, which can effectively mine multi-scale information to alleviate scale variations.

The detailed structure of MSCA is shown in Fig.~\ref{fig:net1}. In MSCA, it is a self-attention based on a two-branch structure. Specifically, the first branch adopts global average pooling (GAP) to exploit global contexts, which can emphasize globally distributed for large objects. The second branch holds the size of the original feature to obtain local contexts, which can prevent small camouflaged objects that are often ignored. Different from several existing multi-scale attention mechanisms, MSCA adopts point convolution in two branches to restore and compress features along the channel dimension, resulting in aggregating multi-scale context information. Cross-level features have different contributions to the COD task, thus fusing multi-level features can complement each other to obtain more comprehensive feature representations. To achieve this, we conduct ACFM on high-level features, and the cross-level fusion process is implemented by
\begin{equation}
\resizebox{.91\linewidth}{!}{$
	\displaystyle
	F_{ab}=\mathcal{M}(F_{a}\uplus F_{b})\otimes F_{a}\oplus(1-\mathcal{M}(F_{a}\uplus F_{b}))\otimes F_{b},
	$}
\end{equation}%
where $\mathcal{M}$ denotes the MSCA operation, $F_{a}$ and $F_{b}$ denote the two input features. Besides, $\oplus$ denotes the initial fusion of $F_{a}$ and $F_{b}$, in which $F_{b}$ is up-sampled and then added with $F_{a}$. $\otimes$ denotes element-wise multiplication. $(1-\mathcal{M}(F_{a}\uplus F_{b}))$ corresponds to the {dotted line} in Fig.~\ref{fig:net1}. Moreover, $F_{ab}$ is then fed into a $3\times{3}$ convolutional layer, followed by batch normalization and a \emph{ReLU} function. Thus, we can obtain the fused cross-level feature, \ie, $F$.

\subsection{Dual-branch Global Context Module}

The proposed ACFM is adopted to fuse multi-scale features, in which a multi-scale channel attention strategy is introduced to obtain informative attention-based features. In addition, global context information is important for boosting the COD performance. Thus, a dual-branch global context module (DGCM) is proposed to fully mine global context information from the fused features. In detail, the output $F\in\mathbb{R}^{{W}\times{H}\times{C}}$ of the ACFM is first fed to two branches with a convolutional layer and average pooling, respectively, thus we obtain the sub-features $F_{c}\in\mathbb{R}^{{W}\times{H}\times{C}}$ and $F_{p}\in\mathbb{R}^{{\frac{W}{2}}\times{\frac{H}{2}}\times{C}}$. After that, in order to learn multi-scale attention-based features, $F_{c}$ and $F_{p}$ are also fed to the MSCA module. Then, an element-wise multiplication is utilized to integrate the feature outputs of MSCA and the corresponding features ($F_{c}$ or $F_{p}$), and then we obtain $F_{cm}\in\mathbb{R}^{{W}\times{H}\times{C}}$ and $F_{pm}\in\mathbb{R}^{{\frac{W}{2}}\times{\frac{H}{2}}\times{C}}$. Further, we adopt an addition operation to integrate the features from the above two-branch structure, and then the fused feature is denoted as $F_{cpm}$. Finally, the residual connection is adopted to combine $F$ with $F_{cpm}$ to obtain ${F}'$. The detailed processing steps are presented as follows:
\begin{equation}
\left\{
\begin{aligned}
&F_{c}= \mathcal{C}\left ( F \right ), F_{cm}= F_{c}\otimes \mathcal{M}\left ( F_{c} \right ),\\
&F_{p}= \mathcal{C}\left( \mathcal{P} \left ( F \right )\right), F_{pm}= F_{p}\otimes \mathcal{M} \left ( F_{p} \right ),\\
&{F}'= \mathcal{C} \left ( F\oplus \mathcal {C} \left ( F_{cm} \oplus \mathcal{U} \left ( F_{pm} \right )\right ) \right ),
\end{aligned}
\right.
\label{eq01}
\end{equation}
where $\mathcal{C}$, $\mathcal{M}$, $\mathcal{P}$, and $\mathcal{U}$ denote the convolution, MSCA, average pooling, and upsampling operation, respectively.

\textbf{Remarks}. It is worth noting that our DGCM aims to enhance the fused features of ACFM, so that the proposed model adaptively extracts multi-scale information from a specific level. For clarity, the overall structure is presented in Fig.~\ref{fig:net1}.

\subsection{Camouflage Inference Module}\label{cim}

To effectively make use of multi-scale features, the low-level features (\ie, $Q_l = \{{f_1,f_2,f_3}\}$) are refined by using the output (denoted as $f_D$) of the second DGCM. The refinement flow is presented to effectively fuse all the features from multiple layers. Followed the setting in the BBS-Net~\cite{BBSNet}, we utilize a modified RFB module (denoted as $\mathcal{R}$) to fully expand the receptive fields for obtaining richer features and reducing the computation. As shown in Fig.~\ref{fig:net1}, $f_D$ is adopted to refine three low-level features by using an element-wise multiplication operation, \ie, $f_1\otimes{f_D}$, $f_2\otimes{f_D}$, and $f_3\otimes{f_D}$. Each refined feature is fed into a RFB module, and we can obtain $\mathcal{R}(f_1\otimes{f_D})$, $\mathcal{R}(f_2\otimes{f_D})$, and $\mathcal{R}(f_3\otimes{f_D})$. Then, $\mathcal{R}(f_1\otimes{f_D})$ and $\mathcal{R}(f_2\otimes{f_D})$ are cascaded and fed into a Conv block. After that, the output and $\mathcal{R}(f_3\otimes{f_D})$ are further cascaded and fed into a Conv block with a channel size of $16$.

Further, we propose a Camouflage Inference Module (CIM) to generate the final detection results, which can take advantage of the low-level features via a multi-scale strategy. In the CIM, to fully exploit multi-scale information, we further propose a Multi-scale Residual Block (MRB) to detect local and multi-scale features. Specifically, we build a two-stream network and each stream utilizes a different convolutional kernel. As shown in Fig.~\ref{fig:net1}, the current feature representation $X$ is fed to the first stream network with two sequential operations, in which the former sequential operation consists of a $3\times{3}$ convolution followed by batch normalization and a \emph{ReLU} function (denoted ``$B_{conv3}$"), and the latter consists of a $3\times{3}$ transpose convolution followed by batch normalization and a \emph{ReLU} function (denoted ``$B_{Dconv3}$"). Besides, $X$ is also fed to the second stream network two other sequential operations, in which the convolutional kernel is $5\times{5}$. The above processing steps can be presented as follows:
\begin{equation}
\left\{
\begin{aligned}
&F_{conv3}= B_{DConv{3\times{3}}}(B_{Conv3\times{3}}(X)),\\
&F_{conv5}= B_{DConv{5\times{5}}}(B_{Conv5\times{5}}(X)).\\
\end{aligned}
\right.
\label{eq01}
\end{equation}

Then, the stream features are cascaded and then fed into a $3\times{3}$ sequential layer to obtain the fused multi-scale feature representation, \ie, $F_{f}=B_{conv3\times{3}}(Cat(F_{conv3}, F_{conv5}))$. To preserve the original information from the input $X$, we adopt residual learning in the MRB. Therefore, we obtain the output of each MRB by adding $X$ and the fused multi-scale feature. The whole CIM consists of three $1\times{1}$ convolution and two MRB, which is used to generate the final prediction $P$.

\subsection{Loss Function}

The binary-cross entropy loss ($\mathcal{L}_\text{BCE}$) is widely used to calculate the loss of each pixel to form a pixel restriction on the network. To effectively exploit the global structure, the work \cite{basnet} introduced IoU loss ($\mathcal{L}_\text{IoU}$) to form a global restriction on the network. However, the above losses are often treated equally to all pixels, ignoring the differences among pixels. To overcome this issue, this work \cite{f3net} improves $\mathcal{L}_\text{BCE}$ and $\mathcal{L}_\text{IoU}$ into the weighted binary cross-entropy loss~($\mathcal{L}_\text{BCE}^w$) and IoU loss ($\mathcal{L}_\text{IoU}^w$). Specifically, by computing the difference between the center pixel and its surroundings, each pixel can be assigned a different weight, so that the hard pixels are paid more attention. In this case, the basic loss function of the proposed model is defined as: $\mathcal{L} = \mathcal{L}_\text{IoU}^w + \mathcal{L}_\text{BCE}^w$. To further achieve the aim of joint optimization, the total loss of our model can be defined by
\begin{align}
& \mathcal{L}= \mathcal{L}(f_D,G)+\mathcal{L}(P,G),
\end{align}%
where $\mathcal{L}(f_D,G)$ denotes the loss between the initial prediction map and ground truth, and $\mathcal{L}(P,G)$ denotes the loss between the final prediction map and ground truth.

\section{Experiments}\label{sec:experiments}

In this section, we first present the implementation details, datasets, and evaluation metrics used in our experiments. Then, we show quantitative and qualitative results by comparing our model with other existing methods. Further, we conduct ablation studies to validate the effectiveness of each key component. Finally, we extend the application of our model to polyp segmentation.
\begin{table*}[th]
	\caption{Quantitative comparison between our model and other state-of-the-art COD methods on three datasets in terms of five metrics (\ie, $S_{\alpha}$, $E_{\phi}$, $F_{\beta}^w$, $F_\beta\uparrow$and $M$).  ``$\downarrow$" / ``$\uparrow$" indicates that smaller or larger is better. Best results are highlighted in \textbf{Bold}.} 
	\resizebox{\textwidth}{!}{
		\renewcommand{\arraystretch}{1.0}
		\setlength\tabcolsep{3.5pt}
		\begin{tabular}{l|l|l|lllll|lllll|lllll}
			\hline\toprule
			\multicolumn{1}{l|}{\multirow{2}{*}{Method}} & 
			\multicolumn{1}{c|}{\multirow{2}{*}{Year}} &\multicolumn{1}{c|}{\multirow{2}{*}{Backbone}}& \multicolumn{5}{c|}{CAMO-Test} & \multicolumn{5}{c|}{CHAMELEON} & \multicolumn{5}{c}{COD10K-Test} \\ \cline{4-18}
			&  &\multicolumn{1}{c|}{}& \multicolumn{1}{c}{$S_\alpha\uparrow$} & \multicolumn{1}{c}{$E_\phi\uparrow$} & \multicolumn{1}{c}{$F_\beta^w\uparrow$} &
			\multicolumn{1}{c}{$F_\beta\uparrow$}& \multicolumn{1}{c|}{$M\downarrow$} & \multicolumn{1}{c}{$S_\alpha\uparrow$} & \multicolumn{1}{c}{$E_\phi\uparrow$} & \multicolumn{1}{c}{$F_\beta^w\uparrow$} & 
			\multicolumn{1}{c}{$F_\beta\uparrow$}&
			\multicolumn{1}{c|}{$M\downarrow$} & \multicolumn{1}{c}{$S_\alpha\uparrow$} & \multicolumn{1}{c}{$E_\phi\uparrow$} & \multicolumn{1}{c}{$F_\beta^w\uparrow$} &
			\multicolumn{1}{c}{$F_\beta\uparrow$}&
			\multicolumn{1}{c}{$M\downarrow$} \\ \midrule
			FPN\cite{fpn}&2017&ResNet-50&0.684&0.791&0.483&0.642&0.131&
			0.794&0.835&0.590&0.676&0.075&
			0.697&0.711&0.411&0.484&0.075\\ 
			MaskRCNN\cite{maskrcnn}&2017&ResNet-50&0.574&0.716&0.43&0.521&0.151&
			0.643&0.780&0.512&0.612&0.099&
			0.613&0.750&0.402&0.470&0.080\\ 
			PSPNet\cite{pspnet}&2017&ResNet-50&0.663&0.778&0.445&0.605&0.139
			&0.773&0.814&0.555&0.650&0.085&
			0.678&0.688&0.377&0.451&0.080\\
			PiCANet\cite{picanet}&2018&ResNet-50&0.609&0.753&0.356&0.538&0.155&
			0.769&0.836&0.536&0.668&0.084&
			0.649&0.678&0.322&0.423&0.083\\
			UNet++\cite{unetplus}&2018&VGGNet-16&0.599&0.740&0.392&0.522&0.149&
			0.695&0.808&0.501&0.586&0.094&
			0.623&0.718&0.350&0.431&0.086\\ \midrule
			BASNet\cite{basnet}&2019&ResNet-34&0.618&0.713&0.413&0.525&0.159&
			0.688&0.742&0.474&0.546&0.118&
			0.634&0.676&0.365&0.421&0.105\\
			CPD\cite{cpd}&2019&ResNet-50&0.716&0.807&0.556&0.675&0.113&
			0.857&0.898&0.731&0.775&0.048&
			0.750&0.792&0.531&0.578&0.053\\
			HTC\cite{htc}&2019&ResNet-50&0.477&0.442&0.174&0.338&0.172&
			0.517&0.490&0.204&0.327&0.129&
			0.548&0.521&0.221&0.298&0.088\\
			MSRCNN\cite{masksrcnn}&2019&ResNet-50&0.618&0.670&0.454&0.544&0.133&
			0.637&0.688&0.443&0.529&0.091&
			0.641&0.708&0.419&0.486&0.073\\
			PoolNet\cite{poolnet}&2019&ResNet-50&0.703&0.790&0.494&0.628&0.128&
			0.776&0.824&0.555&0.649&0.078&
			0.705&0.708&0.416&0.479&0.070\\
			EGNet\cite{egnet}&2019&ResNet-50&0.662&0.780&0.495&0.640&0.125&
			0.750&0.854&0.531&0.694&0.075&
			0.733&0.799&0.519&0.572&0.055 \\ 
			ANet-SRM\cite{anet}&2019&FCN&0.682&0.725&0.484&0.593&0.126&
			\multicolumn{1}{c}{‡} & \multicolumn{1}{c}{‡} & \multicolumn{1}{c}{‡} & \multicolumn{1}{c}{‡} &\multicolumn{1}{c|}{‡}&
			\multicolumn{1}{c}{‡} & \multicolumn{1}{c}{‡} & \multicolumn{1}{c}{‡} & \multicolumn{1}{c}{‡} &\multicolumn{1}{c}{‡} \\ \midrule
			SINet\cite{sinet}&2020&ResNet-50&0.752&0.835&0.606&0.709&0.100&
			0.868&0.899&0.740&0.776&0.044&
			0.771&0.797&0.551&0.593&0.051\\
			MirrorNet\cite{mirrornet}&2020&ResNet-50&0.784&0.849&0.652&0.767&0.078&
			\multicolumn{1}{c}{‡} & \multicolumn{1}{c}{‡} & \multicolumn{1}{c}{‡} & \multicolumn{1}{c}{‡} &\multicolumn{1}{c|}{‡}&
			\multicolumn{1}{c}{‡} & \multicolumn{1}{c}{‡} & \multicolumn{1}{c}{‡} & \multicolumn{1}{c}{‡} &\multicolumn{1}{c}{‡} \\ 
			UCNet\cite{ucnet}&2020&ResNet-50&0.739&0.811&0.640&0.716&0.094&
			0.880&0.929&0.817&0.830&0.036&
			0.776&0.867&0.633&0.673&0.042 \\
			PraNet\cite{pranet}&2020&Res2Net-50&0.769&0.833&0.663&0.715&0.094&
			0.860&0.898&0.763&0.775&0.044&
			0.789&0.839&0.629&0.640&0.045\\ \midrule
			LSR\cite{lsr}&2021&ResNet-50&0.787&0.855&0.696&0.756&0.080&
			0.890&0.936&0.822&0.835&0.030&
			0.804&0.882&0.673&0.699&0.037\\
			PFNet\cite{pfnet}&2021&ResNet-50&0.782&0.852&0.695&0.751&0.085&
			0.882&0.942&0.816&0.820&0.033&
			0.800&0.868&0.660&0.676&0.040\\
			ERRNet\cite{errnet}&2022&ResNet-50
			&0.779 &0.851 &0.679 &0.731 &0.085
			&0.868 &0.917 &0.787 &0.796 &0.039
			&0.786 &0.845 &0.630 &0.646 &0.043\\ 
			\midrule
			\multirow{2}{*}{\ourM}&2022&ResNet-50
			&0.765&0.838&0.680&0.732&0.087
			&0.878&0.938&0.822&0.846&0.031
			&0.776&0.872&0.640&0.688&0.042\\
			{}&2022&Res2Net-50&\textbf{0.800}&\textbf{0.869}&\textbf{0.730}&\textbf{0.777}&\textbf{0.077} &
			\textbf{0.893}&\textbf{0.947}&\textbf{0.845}&\textbf{0.836}&\textbf{0.028}&
			\textbf{0.811}&\textbf{0.891}&\textbf{0.691}&\textbf{0.718}&\textbf{0.036}\\ \bottomrule
		\end{tabular}
	}
	\label{tab1}
\end{table*}
\subsection{Implementation Details}

The proposed model is implemented using PyTorch, and the Res2Net-50~\cite{res2net} pretrained on ImageNet is adopted as the backbone. We adopt the AdaX~\cite{li2020adax} algorithm to optimize the overall framework. To enhance the generalizability of the proposed model, we adopt a multi-scale training strategy $\{0.75, 1, 1.25\}$ in the training stage. Besides, the input images are resized to $352\times{352}$. The initial learning rate is set to $1e-4$, which will decay $10$ times after $30$ epochs. The network is trained on an NVIDIA Tesla V100 GPU. The batch size is set to be $30$, and the training process takes about $6$ hours over $100$ epochs. 

\subsection{Datasets}\label{sec:Datasets}

To validate the effectiveness of the proposed model, we carry out comparison experiments on three benchmark datasets. The details of each COD dataset are provided below.

\begin{itemize}
	\item CHAMELEON~\cite{sinet} is collected using the Google search engine with the keyword ``camouflage animals", which consists of $76$ camouflaged images using as a testing dataset. 
	
	\item CAMO~\cite{anet} consists of $1,250$ camouflaged images ($1,000$ for training, $250$ for testing), which includes eight categories.
	
	\item COD10K~\cite{sinet} is currently the largest COD dataset with pixel-level annotations. It consists of $5,066$ images, where $3,040$ for training and $2,026$ for testing. Besides, this dataset is split into $5$ super-classes and $69$ sub-classes. 
\end{itemize}

\textbf{Training and testing settings}. Following the setting in~\cite{sinet}, for the CAMO dataset, we utilize the default training set. For the COD10K dataset, we utilize the default training camouflaged images. In the test stage, we test our \our~ and other compared methods on the test sets of CAMO and COD10K, and the whole CHAMELEON dataset.

\begin{figure*}[ht]
	\begin{centering}
		\includegraphics[width=\textwidth]{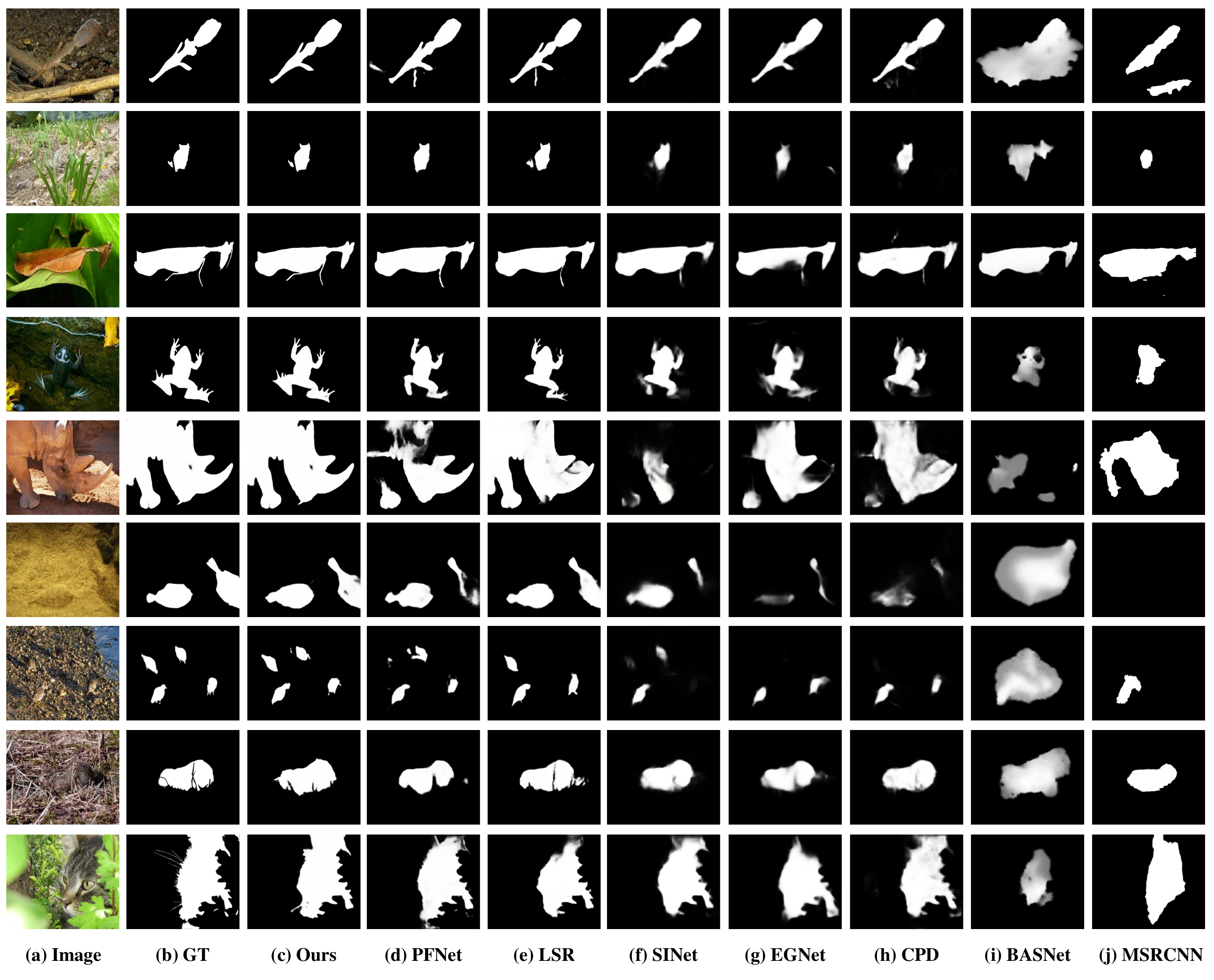}
		\vspace{-5pt}
		\caption{Qualitative results of our model and seven state-of-the-art COD methods (\ie, 
			PFNet~\cite{pfnet}, LSR\cite{lsr}, SINet~\protect\cite{sinet}, EGNet~\protect\cite{egnet}, CPD~\protect\cite{cpd}, BASNet~\protect\cite{basnet}, and MSRCNN~\protect\cite{masksrcnn}).
		}
		\label{results}
	\end{centering}%
\end{figure*}

\subsection{Evaluation Metrics}\label{metrics}

We adopt five widely used metrics, namely MAE ($M$), S-measure ({$S_\alpha$}), F-measure ({$F_\beta$}), weighted F-measure ({$F_\beta^w$}), and E-measure ({$E_\phi$}), to evaluate the effectiveness of our model and all compared methods. The detailed definitions for these metrics are as follows:\par
\begin{enumerate}
	\item Mean absolute error~(MAE, $M$) computes the average difference between the normalized prediction and ground truth, which is defined by
	\begin{equation}
	M = \frac{1}{W\cdot H}\sum_{x=1}^{W} \sum_{y=1}^{H}|P(x,y)-G(x,y)|,
	\end{equation}%
	where $P(x,y)$ and $G(x,y)$ correspond to the prediction map and ground truth value at the pixel location $(x,y)$. $W$ and $H$ are the width and height of the prediction map $P$.
	
	\item S-measure ({$S_\alpha$}) \cite{sm} computes the object-aware similarity $S_o$ and region-aware structure similarity $S_r$ between the prediction map and the ground truth by
	\begin{equation}
	S_\alpha =\alpha * S_o + (1 - \alpha) *S_r,
	\end{equation}%
	where $\alpha$~is a trade-off parameter to balance $S_o$ and $S_r$, and it is set to 0.5 as in \cite{sm}.
	
	\item F-measure ({$F_\beta$}) is employed to evaluate binary map results between the prediction and the ground truth. It is defined as:
	\begin{equation}
	F_\beta = \frac{(1+ \beta^2)Precision \cdot Recall}{\beta^2\ Precision +Recall},
	\end{equation}
	where $\beta$ is the weight between the precision and recall. $\beta^2=0.3$ is the default setting since the precision is often weighted more than the recall. Besides, the improved version of $F_\beta$, \ie, weighted F-measure ({$F_\beta^w$}), is proposed to overcome the interpolation, dependency, and equal-importance flaws of $F_\beta$. This metric synthetically considers both weighted precision and weighted recall~\cite{fm}.
	
	\item E-measure({$E_\phi$})~\cite{em,21Fan_HybridLoss} evaluates the overall and local accuracy of a binary foreground map, which can be defined by
	\begin{equation}
	E_\phi = \frac{1}{W\cdot H}\sum_{x=1}^{W} \sum_{y=1}^{H}\phi(x,y).
	\end{equation}%
\end{enumerate}

In summary, for these five metrics above, higher {$S_\alpha$}, {$F_\beta^w$}, {$F_\beta$}, {$E_\phi$}, and lower $M$ indicate better performance. Note that we adopt adaptive F/E-measure in our evaluation.

\begin{table*}[ht]
	\caption{Results comparison between our model and other state-of-the-art COD methods for the COD10K dataset on four super-classes in terms of four evaluation metrics (\ie, $S_{\alpha}$, $E_{\phi}$, $F_\beta\uparrow$and $M$). ``$\uparrow$" / ``$\downarrow$" indicates that larger or smaller is better. Best results are highlighted in \textbf{Bold} fonts.}
	\resizebox{\textwidth}{!}{
		\renewcommand{\arraystretch}{1.0}
		\setlength\tabcolsep{3.5pt}
		\begin{tabular}{l|l|llll|llll|llll|llll}
			\hline\toprule
			\multicolumn{1}{l|}{\multirow{2}{*}{Method}} & 
			\multicolumn{1}{c|}{\multirow{2}{*}{Year}} & \multicolumn{4}{c|}{Aquatic} & \multicolumn{4}{c|}{Terrestrial} & \multicolumn{4}{c|}{Flying} &\multicolumn{4}{c}{Amphibian} \\ \cline{3-18}
			& \multicolumn{1}{c|}{} & \multicolumn{1}{c}{$S_\alpha\uparrow$} & \multicolumn{1}{c}{$E_\phi\uparrow$} & 
			\multicolumn{1}{c}{$F_\beta\uparrow$}& \multicolumn{1}{c|}{$M\downarrow$} & \multicolumn{1}{c}{$S_\alpha\uparrow$} & \multicolumn{1}{c}{$E_\phi\uparrow$} & 
			\multicolumn{1}{c}{$F_\beta\uparrow$}&
			\multicolumn{1}{c|}{$M\downarrow$} & \multicolumn{1}{c}{$S_\alpha\uparrow$} & \multicolumn{1}{c}{$E_\phi\uparrow$} & 
			\multicolumn{1}{c}{$F_\beta\uparrow$}&
			\multicolumn{1}{c|}{$M\downarrow$}& 
			\multicolumn{1}{c}{$S_\alpha\uparrow$} & \multicolumn{1}{c}{$E_\phi\uparrow$} & 
			\multicolumn{1}{c}{$F_\beta\uparrow$}&
			\multicolumn{1}{c}{$M\downarrow$}\\ \midrule
			FPN\cite{fpn}&2017&0.684&0.730&0.533&0.103&
			0.669&0.679&0.418&0.071&
			0.726& 0.724& 0.500& 0.061&
			0.745&0.772&0.576&0.065 \\
			MaskRCNN\cite{maskrcnn}&2017&0.560&0.721&0.418&0.123&
			0.608&0.749&0.441&0.070&
			0.644&0.767&0.520&0.063&
			0.665&0.785&0.554&0.081\\
			PSPNet\cite{pspnet}&2017&0.659&0.706&0.449&0.111&
			0.658&0.652&0.391&0.074&
			0.700&0.703&0.463&0.067&
			0.736&0.739&0.535&0.072\\
			PiCANet\cite{picanet}&2018&0.628&0.698&0.466&0.115&
			0.625&0.640&0.357&0.075&
			0.677&0.696&0.441&0.069&
			0.704&0.727&0.521&0.079\\
			UNet++\cite{unetplus}&2018&0.599&0.708&0.445&0.121&
			0.593&0.692&0.369&0.081&
			0.659&0.745&0.470&0.068&
			0.677&0.754&0.512&0.079\\ \midrule
			BASNet\cite{basnet}&2019&0.620&0.678&0.451&0.134&
			0.601&0.630&0.351&0.109&
			0.664&0.711&0.454&0.086&
			0.708&0.739&0.529&0.087\\
			CPD\cite{cpd}&2019&0.746&0.806&0.624&0.075&
			0.714&0.757&0.506&0.052&
			0.777&0.809&0.603&0.041&
			0.816&0.854&0.676&0.041\\
			HTC\cite{htc}&2019&0.507&0.495&0.292&0.129&
			0.530&0.485&0.236&0.078&
			0.582&0.559&0.345&0.070&
			0.606&0.598&0.403&0.088\\
			MSRCNN\cite{masksrcnn}&2019&0.614&0.686&0.475&0.107&
			0.610&0.672&0.426&0.070&
			0.675&0.744&0.529&0.058&
			0.722&0.786&0.618&0.055\\
			PoolNet\cite{poolnet}&2019&0.689&0.721&0.519&0.099&
			0.677&0.672&0.417&0.064&
			0.733&0.725&0.496&0.057&
			0.767&0.775&0.583&0.059\\
			EGNet\cite{egnet}&2019&0.693&0.787&0.590&0.088&
			0.711&0.773&0.513&0.049&
			0.771&0.828&0.606&0.039&
			0.787&0.835&0.649&0.048\\ \midrule
			SINet\cite{sinet}&2020&0.758&0.806&0.631&0.073&
			0.743&0.765&0.532&0.050&
			0.798&0.817&0.614&0.040&
			0.827&0.847&0.684&0.042\\
			UCNet\cite{ucnet}&2020&0.768&0.853&0.701&0.060&
			0.742&0.845&0.608&0.042&
			0.807&0.892&0.705&0.030&
			0.827&0.916&0.751&0.034\\ \midrule
			PraNet\cite{pranet}&2020&0.780&0.832&0.670&0.065&
			0.756&0.810&0.577&0.046&
			0.819&0.864&0.668&0.033&
			0.842&0.892&0.730&0.035\\
			LSR\cite{lsr}&2021&0.803&0.873&0.727&0.052&
			0.772&0.858&0.639&\textbf{0.038}&
			0.830&0.907&0.726&\textbf{0.026}&
			0.846&\textbf{0.922}&0.774&\textbf{0.030}\\
			PFNet\cite{pfnet}&2021&0.793&0.865&0.706&0.056&
			0.773&0.850&0.621&0.041&
			0.824&0.887&0.698&0.030&
			\textbf{0.848}&0.896&0.753&0.031\\
			ERRNet\cite{errnet}&2022&0.781&0.846&0.670&0.061&
			0.753&0.816&0.580&0.045&
			0.814&0.867&0.676&0.032&
			0.834&0.885&0.736&0.036\\ \midrule
			\ourM&2022&\textbf{0.810}&\textbf{0.884}&\textbf{0.736}&\textbf{0.049}&
			\textbf{0.782}&\textbf{0.869}&\textbf{0.665}&0.039&
			\textbf{0.835}&\textbf{0.915}&\textbf{0.750}&\textbf{0.026}&
			\textbf{0.848}&0.920&\textbf{0.784}&0.031\\			
			\bottomrule			
		\end{tabular}
	}
	\label{tab2}
\end{table*}
\begin{figure*}[hb]
	\centering{
		\includegraphics[width=0.95\linewidth]{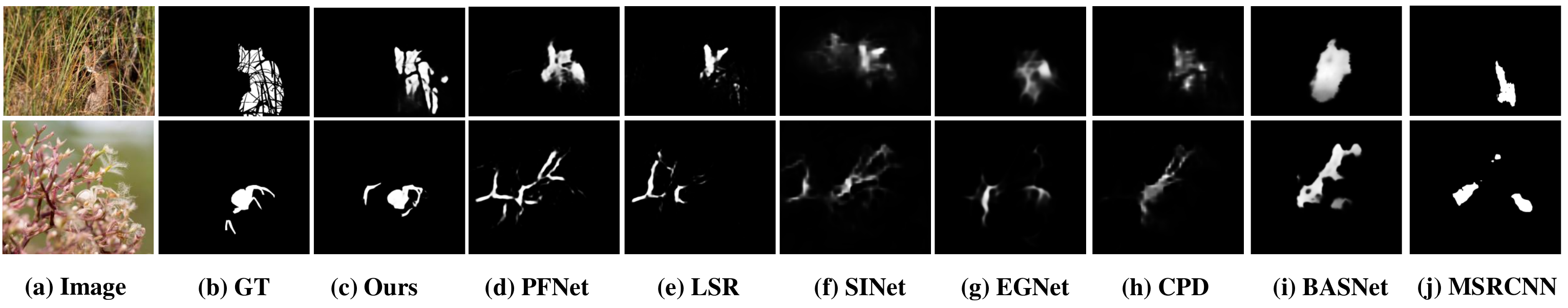}
		\vspace{-5pt}
		\caption{Some failure cases of our model and seven comparison methods.}
		\label{fig:failure}}
\end{figure*}

\subsection{Comparison with State-of-the-art COD Methods}\label{comparison}

We compare the proposed method with 19 state-of-the-art (SOTA) COD methods, including FPN~\cite{fpn}, MaskRCNN~\cite{maskrcnn}, PSPNet~\cite{pspnet}, UNet++~\cite{unetplus}, PiCANet~\cite{picanet}, MSRCNN~\cite{masksrcnn}, BASNet~\cite{basnet}, PFANet~\cite{pfanet}, CPD~\cite{cpd}, HTC~\cite{htc}, EGNet~\cite{egnet}, ANet-SRM~\cite{anet}, SINet~\cite{sinet}, MirrorNet~\cite{mirrornet}, UCNet~\cite{ucnet}, PraNet~\cite{pranet}, LSR~\cite{lsr}, PFNet~\cite{pfnet}, and ERRNet~\cite{errnet}.
For a fair comparison, the results of MirrorNet are collected from~link\footnote{\url{https://sites.google.com/view/ltnghia/research/camo}}, the results of ERRNet are taken from~link\footnote{\url{https://github.com/GewelsJI/ERRNet}}, and the results of remaining 18 methods are collected from link\footnote{\url{http://dpfan.net/socbenchmark}}.

\subsubsection{Quantitative Evaluation}

Table~\ref{tab1} shows the quantitative results of the proposed model, all compared COD methods on the three benchmark datasets. Meanwhile, the backbones of all the methods are provided in Table~\ref{tab1}. 

\textbf{Performance on CAMO.} From Table~\ref{tab1}, compared with 19 SOTA models, it can be observed that our \ourM~performs better than other models in terms of all metrics. Specifically, compared with LSR, the second-best model, $S_\alpha$, $F_\beta^w$, and $E_\phi$ increased by $1.6\%$, $4.9\%$, and $1.6\%$, respectively. This is mainly because that the proposed ACFM can mine multi-scale information to alleviate scale variations, and DCCM aims to enhance the fused features of ACFM for providing global context information. More importantly, the proposed CIM can make use of the low-level features and further exploit multi-scale information to boost the COD performance. 

\textbf{Performance on CHAMELEON.} Our \ourM~is also tested on CHAMELEON \cite{sinet} dataset. As reported in Table~\ref{tab1}, our \ourM~obtains better performances across all the metrics. Specifically, compared with the LSR, $E_\phi$ and $F_\beta^w$ increase by $1.2\%$ and $2.7\%$, respectively. Overall, the results demonstrate the good generalization ability of our model. \par

\textbf{Performance on COD10K.} The COD10K dataset is a more challenging sequence, which contains $2,026$ testing images distributed in $5$ super-classes. It can be observed that our \ourM~achieves the new SOTA performances across all five metrics. The main reason for this robustness results is that the mature designed modules can combine multi-level features and exploit rich global context information.

\begin{table*}[h]
	\caption{Ablation studies on the three datasets. Best results are highlighted in \textbf{Bold}. Ver. =  Version.} \vspace{-0.22cm}
	\resizebox{0.95\textwidth}{!}{
		\renewcommand{\arraystretch}{1.}
		\setlength\tabcolsep{2.5pt}
		\begin{tabular}{l|l|lllll|lllll|lllll}
			\toprule
			\multirow{2}{*}{Ver.} & \multicolumn{1}{l|}{\multirow{2}{*}{Method}} & \multicolumn{5}{c|}{CAMO-Test} & \multicolumn{5}{c|}{CHAMELEON} & \multicolumn{5}{c}{COD10K-Test} \\ \cline{3-17} 
			& \multicolumn{1}{c|}{} & \multicolumn{1}{c}{$S_\alpha\uparrow$} & \multicolumn{1}{c}{$E_\phi\uparrow$} & \multicolumn{1}{c}{$F_\beta^w\uparrow$} &
			\multicolumn{1}{c}{$F_\beta\uparrow$}&
			\multicolumn{1}{c|}{$M\downarrow$} & \multicolumn{1}{c}{$S_\alpha\uparrow$} & \multicolumn{1}{c}{$E_\phi\uparrow$} & \multicolumn{1}{c}{$F_\beta^w\uparrow$} &
			\multicolumn{1}{c}{$F_\beta\uparrow$}&
			\multicolumn{1}{c|}{$M\downarrow$} & \multicolumn{1}{c}{$S_\alpha\uparrow$} & \multicolumn{1}{c}{$E_\phi\uparrow$} & \multicolumn{1}{c}{$F_\beta^w\uparrow$} &
			\multicolumn{1}{c}{$F_\beta\uparrow$}&
			\multicolumn{1}{c}{$M\downarrow$} \\ \midrule
			No.1 & Basic&0.773&0.843&0.684&0.742&0.090&
			0.883&0.923&0.813&0.823&0.033&
			0.805&0.870&0.678&0.683&0.038
			\\ \midrule
			No.2 & Basic+ACFM & 0.784&0.852&0.699&0.751&0.087&
			0.885&0.932&0.821&0.829&0.032&
			0.810&0.881&0.678&0.697&\textbf{0.036}\\ \midrule
			No.3 & Basic+DGCM & 0.786&0.856&0.696&0.752&0.081&
			0.883&0.940&0.820&0.834&0.030& 
			0.808&0.881&0.676&0.695&0.037\\ \midrule
			No.4 & Basic+\ourB &0.787&0.857&0.700&0.747&0.086&
			0.885&0.929&0.820&0.827&0.032&
			0.805&0.873&0.669&0.684&0.038 \\ \midrule
			No.5 & Basic+ACFM+DGCM &0.796&0.864&0.719&0.764&0.080&
			0.888&0.932&0.828&\textbf{0.844}&0.032&
			\textbf{0.813}&0.886&0.686&0.703&\textbf{0.036}\\ \midrule
			No.6 &{MSCA} $\rightarrow$ {Conv}&
			0.785&0.852&0.700&0.748&0.084&
			0.888&0.935&0.826&0.835&0.031&
			0.810&0.873&0.679&0.693 &0.037  \\ \midrule
			No.7 & Basic+ACFM+DGCM+\ourB &\textbf{0.800}&\textbf{0.869}&\textbf{0.730}&\textbf{0.777}&\textbf{0.077} &
			\textbf{0.893}&\textbf{0.947}&\textbf{0.845}&{0.836}&\textbf{0.028}&
			0.811&\textbf{0.891}&\textbf{0.691}&\textbf{0.718}&\textbf{0.036}\\ \bottomrule
		\end{tabular}
	}
	\label{tab3}
\end{table*}

\begin{figure*}[ht]
	\centering{
		\includegraphics[width=0.92\linewidth]{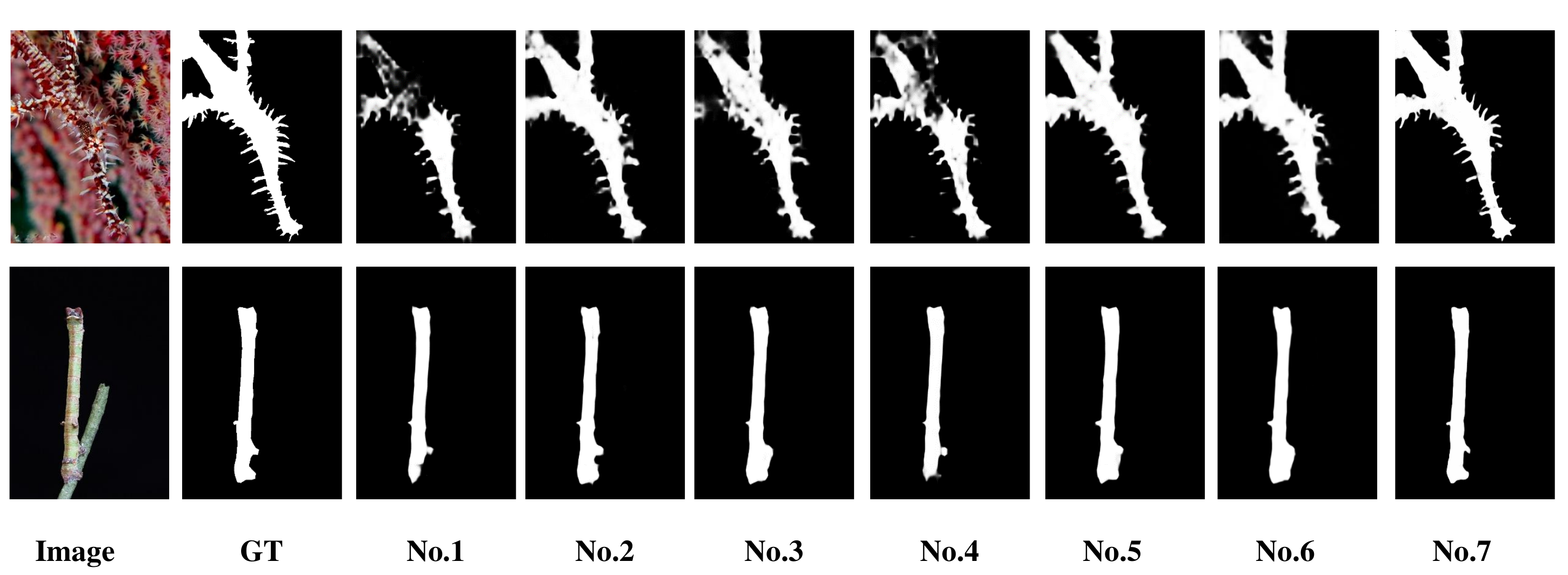}
		\vspace{-5pt}
		\caption{Visual comparisons for showing the benefits of different modules. The results of No.1$\sim$No.7 are shown in Table~\ref{tab3}.}
		\label{fig:ablation}}
\end{figure*}
\subsubsection{Per-subclass Performance}\label{subclass}

In addition to the overall quantitative comparison on the COD10K dataset, the super-class results are also presented in Table~\ref{tab2}. In this experiment, four main super-classes, including ``Aquatic", ``Terrestrial", ``Flying", and ``Amphibian", are taken into consideration. 
From the results in Table~\ref{tab2}, it can be seen that \ourM~outperforms other SOTA methods on the super-class ``Aquatic" and ``Flying" in terms of all four metrics.
On the super-class ``Terrestrial", as shown in Table~\ref{tab2}, our model achieves the best performance in terms of {$F_\beta$}, {$E_\phi$}, and {$S_\alpha$} metrics. Except for {$E_\phi$} and $MAE$, the proposed method outperforms other SOTA models on the super-class ``Amphibian". 

\subsubsection{Qualitative Evaluation}

\figref{results} shows the qualitative comparison results. These examples are collected from five super-classes in the COD10K dataset, including ``Aquatic", ``Terrestrial", ``Flying", and ``Amphibian", and ``Others'' (from 1$^{st}$ row to 5$^{th}$ row). The challenge factors are also presented, \ie, multiple objects (see the 6$^{th}$ and 7$^{th}$ rows) and occlusion (see the 8$^{th}$ and 9$^{th}$ rows). 
As shown in \figref{results}, \ourM~improves the visual results compared to other COD methods in the detailed branches of the ghost pipefish and mantis (see the 1$^{st}$ and 3$^{rd}$ rows). In the context under different lightning, compared with PFNet \cite{pfnet}, LSR \cite{lsr}, and SINet \cite{sinet}, our model can accurately identify camouflaged objects and their boundary (see the 5$^{th}$ row). For multiple objects, compared with PFNet \cite{pfnet}, LSR \cite{lsr}, \ourM~ effectively detect multiple concealed objects (see the 6$^{th}$ and 7$^{th}$ rows). To detect occluded objects, compared with other methods, our model can recognize the certain boundaries of the camouflaged objects. Overall, compared with other SOTA models, our \our~can achieve better performance by detecting more accurate and complete camouflaged objects with rich details.

\subsubsection{Model Size and Inference Time}
The results, shown in Tables~\ref{tab:model_size}, are the model sizes and inference times of our model and representative models. \# Param is measured in million (M) and memory is measured in giga (G).
As can be observed, our model is with the minimal parameters in comparison with the representative models. Its memory cost is also controlled well, \ie, much less than ERRNet and PFNet.
However, we realize that the inference speed of our \ourM~is not promising enough, which is mainly caused by the new improvement applied to our model. Specifically, the IJCAI version of our model can achieve an Fps of 39.07, which is reduced to 27.41 after using the CIM. We will seek to improve the CIM with more efficient design in future work.
Note that we perform experiments to test the inference time on a single Tesla P100 GPU.

\begin{table}[t]
	\centering
	\small
	\caption{\small Comparison of model size and inference time.
	}\label{tab:model_size}
	\renewcommand{\arraystretch}{1.2}
	\setlength{\tabcolsep}{5pt}
	
	\resizebox{0.42\textwidth}{!}{\begin{tabular}{ll||c|c|c}
		\hline
		&Method      & \# Param (M)   &Memory (G) &Speed (Fps)        \\
		\hline
		&SINet\cite{sinet}&48.946 &19.475& 30.85 \\
		&PraNet\cite{pranet}&30.498&13.078&42.60 \\
		&LSR\cite{lsr}&28.720&17.423&37.96\\
		&PFNet\cite{pfnet}&18.977 &46.498&38.84  \\
		&ERRNet\cite{errnet}&67.708 &40.050&53.34 \\
		&C$^2$F-Net&18.063 &25.214&27.41 \\
		\hline
	\end{tabular}}
\end{table}

\subsubsection{Failure Cases}
We present a number of typical failure cases in \figref{fig:failure}.
As can be observed, our model may fail in some extremely challenging cases. However, it is worth noting that, in these cases, existing SOTA models also fail and our model still outperforms the SOTA models.
\begin{table*}[ht]
	\caption{Quantitative evaluation for Polyp segmentation on the four datasets. The best results are highlighted in \textbf{Bold} fonts.} \vspace{-0.25cm}
	\resizebox{\textwidth}{!}{
		\renewcommand{\arraystretch}{1.1}
		\setlength\tabcolsep{1.5pt}
		
		\begin{tabular}{l|lllll|lllll|lllll|lllll}
			\hline \toprule
			\multicolumn{1}{l|}{\multirow{2}{*}{Method}} & \multicolumn{5}{c|}{CVC-300} &
			\multicolumn{5}{c|}{CVC-ClinicDB} & \multicolumn{5}{c|}{CVC-ColonDB} &
			\multicolumn{5}{c}{ETIS-LaribPolupDB}\\ \cline{2-21}
			\multicolumn{1}{c|}{}&
			\multicolumn{1}{c}{$S_\alpha\uparrow$} & \multicolumn{1}{c}{$E_\phi\uparrow$} & \multicolumn{1}{c}{$F_\beta^w\uparrow$} &
			\multicolumn{1}{c}{$F_\beta\uparrow$}&
			\multicolumn{1}{c|}{$M\downarrow$} &
			
			\multicolumn{1}{c}{$S_\alpha\uparrow$} & \multicolumn{1}{c}{$E_\phi\uparrow$} & \multicolumn{1}{c}{$F_\beta^w\uparrow$} &
			\multicolumn{1}{c}{$F_\beta\uparrow$} &
			\multicolumn{1}{c|}{$M\downarrow$} &
			
			\multicolumn{1}{c}{$S_\alpha\uparrow$} & \multicolumn{1}{c}{$E_\phi\uparrow$} & \multicolumn{1}{c}{$F_\beta^w\uparrow$} &
			\multicolumn{1}{c}{$F_\beta\uparrow$} &
			\multicolumn{1}{c|}{$M\downarrow$} &
			\multicolumn{1}{c}{$S_\alpha\uparrow$} & \multicolumn{1}{c}{$E_\phi\uparrow$} & \multicolumn{1}{c}{$F_\beta^w\uparrow$} &
			\multicolumn{1}{c}{$F_\beta\uparrow$} &
			\multicolumn{1}{c}{$M\downarrow$} \\ \midrule
			UNet\cite{unet}&0.843 &0.867 &0.684& 0.703& 0.022&
			0.889& 0.917&0.811& 0.804&0.019& 
			0.712&0.763&0.498&0.569&0.061&
			0.684&0.645&0.366&0.398&0.036\\ \midrule
			UNet++\cite{unetplus}&0.839&0.884 &0.687&0.706  & 0.018&
			0.873& 0.898&0.785& 0.784& 0.022&
			0.691&0.762&0.467&0.560&0.064&
			0.683&0.704&0.390&0.465&0.035\\ \midrule
			SFA\cite{fang2019selective}&0.640& 0.604& 0.341 &0.353 & 0.065&
			0.793&0.816 &0.647&0.655 &0.042&
			0.634&0.648&0.379&0.407&0.094&
			0.557&0.515&0.231&0.255&0.109\\ \midrule
			PraNet\cite{pranet}&0.925& 0.938&0.843 &0.824 &0.010&
			0.936& 0.957&0.896&0.885 &\textbf{0.009}&
			0.820&0.847&0.699&0.718&\textbf{0.043}&
			0.794&0.792&0.600&0.602&\textbf{0.031}\\ \midrule
			\ourM& \textbf{0.928}&\textbf{0.970} &\textbf{0.864}&\textbf{0.868 } &\textbf{0.008 }
			& \textbf{0.942}&\textbf{0.978}& \textbf{0.923}&\textbf{0.922}& \textbf{0.009}&
			\textbf{0.821}&\textbf{0.883}&\textbf{0.713}&\textbf{0.735}&0.044&
			\textbf{0.798}&\textbf{0.853}&\textbf{0.629}&\textbf{0.649}&0.032\\ \bottomrule
			
		\end{tabular}
	}
	\label{tab4}
\end{table*}

\begin{figure*}[hb]
	\centering{
		\includegraphics[width=0.95\linewidth]{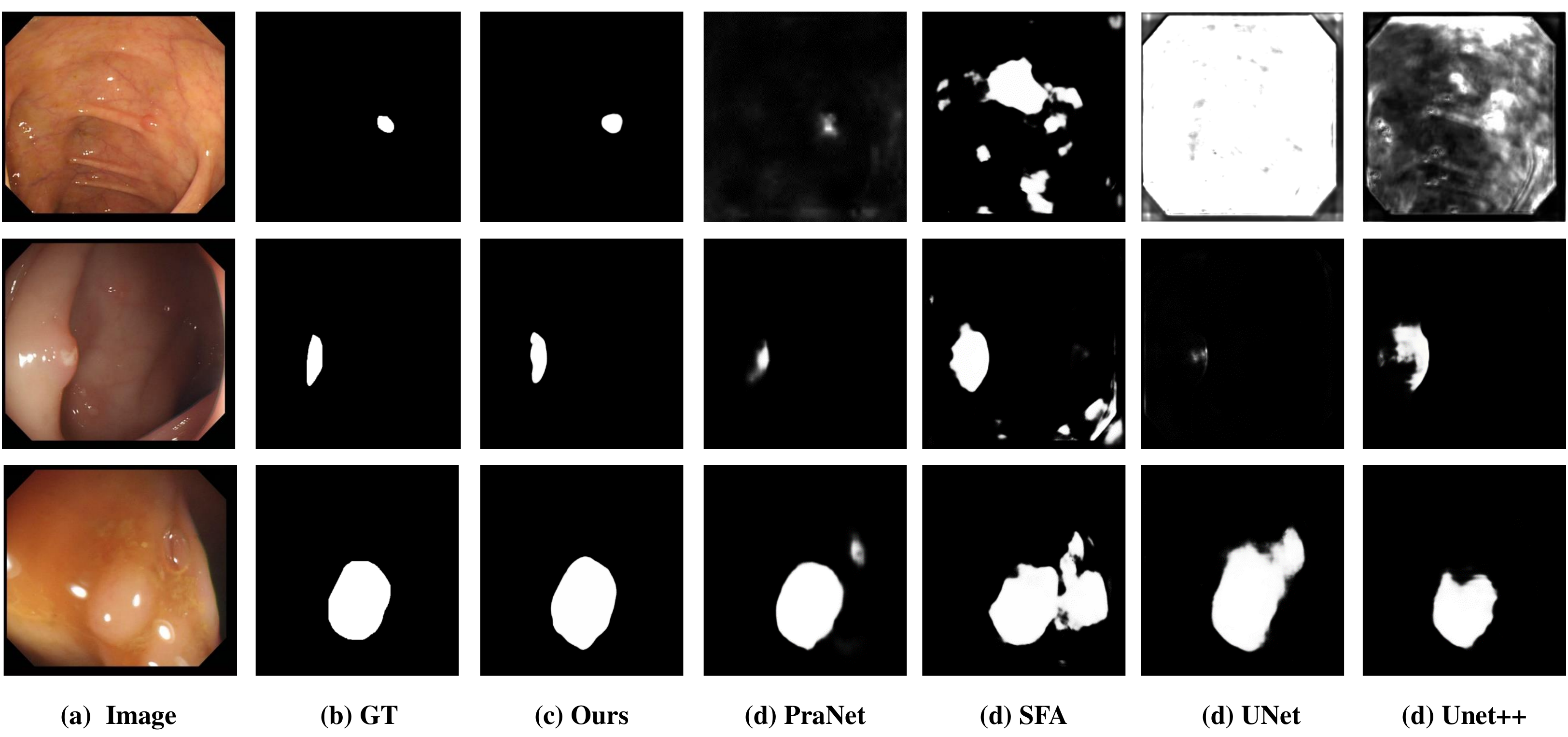}
		\vspace{-5pt}
		\caption{Visual comparisons of different methods on the polyp segmentation task.}
		\label{fig:polyp}
	}
\end{figure*}

\subsection{Ablation Study}

To evaluate the effectiveness of each component in the proposed model, we designed seven ablation experiments (as presented in Table~\ref{tab3}). In the No.1 (Basic) experiment, we removed all ACFMs, DGCMs, and \ourB, while keeping RFBs. The features from RFBs are added for predicting the COD result.
In the No.2 (Basic+ACFM) experiment, we directly connected the two ACFMs by removing DGCMs and \ourB. In the No.3 (Basic+DGCM) experiment, we replaced ACFMs with the combined operation of upsampling and then conducted an addition operation, while DGCMs keeping unchanged and \ourB~was removed. In the No.4 (Basic+\ourB) experiment, we refined the features and then predicted the COD result with our \ourB.
In the No.5 (Basic+ACFM+DGCM) experiment, ACFMs and DGCMs are kept, while the \ourB~is taken out. In the No.6 ({MSCA} $\rightarrow$ {Conv}) experiment, we replaced MSCA with a convolutional block. Finally, No.7 (Basic+ACFM+DGCM+\ourB) is the full version of our \ourB. \par

\textbf{Effectiveness of ACFM}. We first study the effectiveness of ACFM. As shown in Table~\ref{tab3}, it can be observed that No.2 (Basic+ACFM) outperforms No.1 (Basic), clearly indicating that the ACFM is helpful and critical to boost the COD performance, increasing the $F_\beta^w$ by $2.15\%$ on CAMO. Compared to the basic network, the ACFM could improve the ability to detect the main part of the camouflaged object (4$^{th}$col).

\textbf{Effectiveness of DGCM}. From the results in Table~\ref{tab3}, it can be seen that No.3 (Basic + DGCM) outperforms better than No.1 (Basic) on the all used datasets. Specifically, the adaptive $E_\phi$ increased  $1.81\%$ on CHAMELEON. The results indicate that introducing the DGCM can enable our model to detect objects accurately.

\textbf{Effectiveness of \ourB}. As shown in Table~\ref{tab3}, No.4 (Basic + CIM)~outperforms No.1 (Basic) on three benchmarks datasets. This owes to CIM's ability of exploiting multi-scale information from low-level features, which plays a crucial role in boosting the COD performance.

\textbf{Effectiveness of ACFM \& DGCM}. To evaluate the combination of ACFM and DGCM, we carry out the ablation study of No.5. As shown in Table~\ref{tab3}, the performance of the combination is generally better than the first four settings.

\textbf{Effectiveness of MSCA}. In this ablation study, we replace it with a convolutional layer (denoted ``MSCA$\rightarrow$Conv''). The comparison results of the No.6 are shown in Table~\ref{tab3}, indicating that the use of MSCA significantly improves the results, especially on the CAMO-Test dataset.

\textbf{Effectiveness of \ourB~\& ACFM \& DGCM}. To evaluate the full version of our model, we test the performance of No.7. From the results in Table~\ref{tab3}, it can be seen that the full version of \ourM~is better than all other ablated ones. Additionally, the visual comparisons in Fig.~\ref{fig:ablation} further show that our full model is more conducive to identifying and locating camouflaged objects.

It is worth noting that No.5 (Basic+ACFM+DGCM) is the conference version of our model, which was published in IJCAI 2021 \cite{c2f}. The results, shown in Table~\ref{tab1}, indicate that the full version of our \ourM~(\ie, No.7) outperforms No. 5 on all three datasets. This sufficiently demonstrates the effectiveness of our new improvement applied to the model.
The underlying reason for effectiveness lies in the ability of CIM in exploiting useful information from low-level features with the guidance of coarse prediction map. In addition, the MRBs in CIM promote making full use of the features for improved performance. Due to these novel designs, CIM effectively improves the performance and provides results much better than the previous version~\cite{c2f}.

\subsection{Application to Polyp Segmentation}\label{polyseg}

As aforementioned in Sec.~\ref{applications}, COD has rich downstream applications in practice. Therefore, we further evaluate our \ourM~by applying it to a typical COD downstream application, polyp segmentation.

Experiments are conducted on four public datasets, including ETIS \cite{ETIS}, CVC-ClinicCB \cite{clinicDB}, CVC-ColonDB \cite{colonDB}, and CVC-300 \cite{CVC-300}. We compare \ourM~with four cutting-edge polyp segmentation models, \ie, UNet \cite{unet}, UNet++ \cite{unetplus}, SFA \cite{fang2019selective}, and PraNet \cite{pranet}. The results of these four methods are taken from \url{https://github.mscom/DengPingFan/PraNet}.
Similar to the COD experiments, we evaluate the results using five metrics, including MAE, {$S_\alpha$}, {$F_\beta^w$}, {$F_\beta$}, and {$E_\phi$}.

\subsubsection{Qualitative Evaluation}
Table~\ref{tab4} summarizes the quantitative results of different methods on four polyp datasets. 

\textbf{Performance on CVC-300}. As shown in Table~\ref{tab4}, \ourM~outperforms the other four cutting-edge methods on CVC-300. Compared with the second-best method, ParNet \cite{pranet}, the performance of {$E_\phi$} improves by $3.30\%$.

\textbf{Performance on CVC-ClinicDB}. We further test our model on CVC-ClinicDB, which includes $612$ open-access images from 31 colonoscopy clips. The performance of {$F_\beta$} increases by $4.01\%$. 

\textbf{Performance on CVC-ColonDB}. CVC-ColonDB is a small dataset, which contains $380$ images from 15 short colons copy sequences. Except for the MAE, \ourM~obtains better performance. Especially, compared to ParNet~\cite{pranet}, the performance of {$E_\phi$} increases by $4.01\%$.

\textbf{Performance on ETIS}. ETIS consists of $196$ polyp images for early diagnosis of colorectal cancer. Compared to ParNet~\cite{pranet}, the results of {$E_\phi$} and {$F_\beta$} are increased by $7.15\%$ and $7.24\%$, respectively. 

\subsubsection{Qualitative Evaluation}
Fig.~\ref{fig:polyp} shows the visual results of different models on four polyp datasets, including small polyps (1$^{st}$ row), different lightning (2$^{nd}$ row), and low-contrast scenes (3$^{rd}$ row). For the small polyps, \ourM~can accurately detect boundaries, while UNet~\cite{unet} and UNet++~\cite{unetplus} fail.
Furthermore, our model outperforms competing methods under different lightning (2$^{nd}$ row), demonstrating its ability to identify polyps under poor visual conditions.
In the condition of low-contrast scenes, the results of \ourM~make the minimum error areas (3$^{rd}$ row). We can see that \ourM~performs better than PraNet \cite{pranet}, the SOTA poly segmentation model that is designed to handle challenging conditions, such as varied size, complicated context, and so on.

\section{Conclusion}\label{conclusion}

In this paper, we have proposed a novel \ourM~for the COD task. Our \ourM~can effectively integrate the cross-level features that contain rich global context information. In \ourM, DGCMs are proposed to exploit global context information from the fused features. Furthermore, the high-level features are fused with ACFMs, which integrate the features under the guidance of valuable attention cues provided by MSCAs.
Finally, we propose CIM to predict the final result from the low-level features enhanced by the coarse prediction map.
Experiments on three public datasets show that our model outperforms other state-of-the-art COD methods. Moreover, the ablation studies validate the effectiveness of each key component. Further evaluation on the four polyp segmentation datasets demonstrates the promising potentials of our \ourM~in COD downstream applications.

\section*{Acknowledgment}
This work was supported in part by the National Science Fund of China under Grant 62172228, an Open Project of the Key Laboratory of System Control and Information Processing, Ministry of Education (Shanghai Jiao Tong University, ID: Scip202102), and the Fundamental Research Funds for the Central Universities (Grant No. D5000220213).

{
	\bibliographystyle{IEEEtranS.bst}
	\bibliography{references}
}


 
 \begin{IEEEbiography}[{\includegraphics[width=1in,height=1.25in,clip,keepaspectratio]{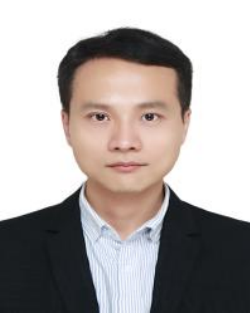}}]{Geng Chen}
 	is a Professor at Northwestern Polytechnical University, China. He received his Ph.D. from Northwestern Polytechnical University, China, in 2016. He was a research scientist at the Inception Institute of Artificial Intelligence, UAE, from 2019 to 2021, and a postdoctoral research associate at the University of North Carolina at Chapel Hill, USA, from 2016 to 2019.
 	His research interests lie in medical image analysis and computer vision.
 \end{IEEEbiography}
 \begin{IEEEbiography}[{\includegraphics[width=1in,height=1.25in,clip,keepaspectratio]{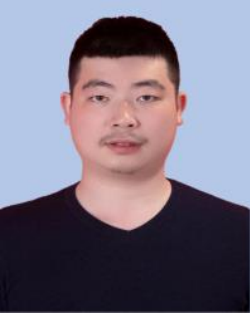}}]{Si-Jie Liu}
	is currently a Ph.D. student with School of Power and Energy, Northwestern Polytechnical University, China, 
	under the supervision of Prof Ya-Feng Wu. 
	His research interests mainly focus on computer measurement \& control, digital twin for industrial equipment maintenance, and computer vision.
\end{IEEEbiography}
 \begin{IEEEbiography}[{\includegraphics[width=1in,height=1.25in,clip,keepaspectratio]{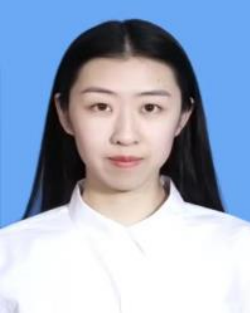}}]{Yu-Jia Sun}
	is currently a master student with School of Computer Science, Inner Mongolia university, China, under the supervision of Prof Hong-Xi Wei.
	Her research interests mainly focus on computer vision and camouflaged object detection.
\end{IEEEbiography}

 \begin{IEEEbiography}[{\includegraphics[width=1in,height=1.25in,clip,keepaspectratio]{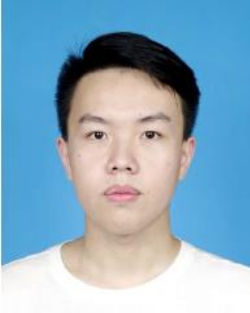}}]{Ge-Peng Ji}
	is currently a master student of Communication
	and Information System at School of Computer
	Science, Wuhan University. His research interests lie in designing deep neural networks and applying deep learning in various fields of lowlevel vision, such as RGB salient object detection, RGB-D salient object detection, video object segmentation, concealed object detection, and medical image segmentation.
\end{IEEEbiography}
 \begin{IEEEbiography}[{\includegraphics[width=1in,height=1.25in,clip,keepaspectratio]{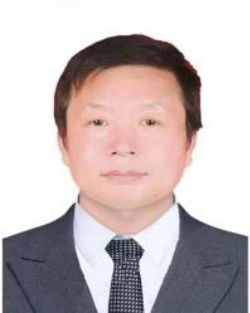}}]{Ya-Feng Wu}
	is a Professor with School of Power and Energy, Northwestern Polytechnical University, China. He received his Ph.D. from Northwestern Polytechnical University, China, in 2002.
	His main research interests include modern signal processing, active noise control, and computer measurement \& control.
\end{IEEEbiography}

 \begin{IEEEbiography}[{\includegraphics[width=1in,height=1.25in,clip,keepaspectratio]{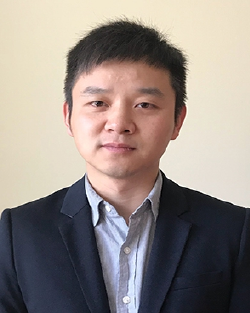}}]{Tao Zhou}
	received the Ph.D. degree in Pattern Recognition and Intelligent Systems from the Institute of Image Processing and Pattern Recognition, Shanghai Jiao Tong University, in 2016. From 2016 to 2018, he was a Postdoctoral Fellow in the BRIC and IDEA lab, University of North Carolina at Chapel Hill. From 2018 to 2020, he was a Research Scientist at the Inception Institute of Artificial Intelligence (IIAI), Abu Dhabi. He is currently a Professor in the School of Computer Science and Engineering, Nanjing University of Science and Technology, China. His research interests include machine learning, computer vision, and medical image analysis.
\end{IEEEbiography}


\end{document}